\documentclass[graybox]{svmult}

\usepackage{type1cm}        
\usepackage{makeidx}         
\usepackage{graphicx}        
\usepackage{multicol}        
\usepackage[bottom]{footmisc}

\usepackage{newtxtext}       %
\usepackage{newtxmath}       

\usepackage{caption, subcaption} 

\makeindex             

\begin{document}

\title*{Face  Recognition using 3D CNNs}
\author{Nayaneesh Kumar Mishra and Satish Kumar Singh}
\institute{Nayaneesh Kumar Mishra \at Computer Vision and Biometric Lab, IIIT Allahabad, India \email{nayaneesh@gmail.com}
\and Satish Kumar Singh \at Computer Vision and Biometric Lab, IIIT Allahabad, India \email{sk.singh@iiita.ac.in}}
%
%
\maketitle

\abstract*{The area of face recognition is one of the most widely researched areas in the domain of computer vision and biometric. This is because, the non-intrusive nature of face biometric makes it comparatively more suitable for application in area of surveillance at public places such as airports. The application of primitive methods in face recognition could not give very satisfactory performance. However, with the advent of machine and deep learning methods and their application in face recognition, several major breakthroughs were obtained. The use of 2D Convolution Neural networks(2D CNN) in face recognition crossed the human face recognition accuracy and reached to 99\%. Still, robust face recognition in the presence of real world conditions such as variation in resolution, illumination and pose is a major challenge for researchers in face recognition. In this work, we used video as input to the 3D CNN architectures for capturing both spatial and time domain information from the video for face recognition in  real world environment. For the purpose of experimentation, we have developed our own video dataset called CVBL video dataset. The use of 3D CNN for face recognition in videos shows promising results with DenseNets performing the best with an accuracy of 97\% on CVBL dataset. }

\abstract{The area of face recognition is one of the most widely researched areas in the domain of computer vision and biometric. This is because, the non-intrusive nature of face biometric makes it comparatively more suitable for application in area of surveillance at public places such as airports. The application of primitive methods in face recognition could not give very satisfactory performance. However, with the advent of machine and deep learning methods and their application in face recognition, several major breakthroughs were obtained. The use of 2D Convolution Neural networks(2D CNN) in face recognition crossed the human face recognition accuracy and reached to 99\%. Still, robust face recognition in the presence of real world conditions such as variation in resolution, illumination and pose is a major challenge for researchers in face recognition. In this work, we used video as input to the 3D CNN architectures for capturing both spatial and time domain information from the video for face recognition in  real world environment. For the purpose of experimentation, we have developed our own video dataset called CVBL video dataset. The use of 3D CNN for face recognition in videos shows promising results with DenseNets performing the best with an accuracy of 97\% on CVBL dataset. 
}

\section{INTRODUCTION}

Face recognition started long back in the 1990s and since then the algorithms have become more efficient. Various
algorithms were applied to detect face in an image and subsequently the recognition of face was done using a recognition algorithm. Researchers developed various mathematical models and features to represent and recognise faces. The features were based on traits of face such as geometry, texture, color and appearance ~\cite{eigenfaces}, ~\cite{features_templates}, ~\cite{human_machine_faces} ~\cite{elastic_faces}, ~\cite{statistical_pr}, ~\cite{viola} .  No feature was able to represent the face with all its complex dimensions. In addition to this, recognition of face was made difficult by real world challenges such as varying illumination, pose, resolution. Various image transformations, super-resolution methods have been proposed to deal with these challenges \cite{face_lbp}, ~\cite{face_quantization}, ~\cite{eigenface_super-resolution}, ~\cite{face_super-resolution}, ~\cite{face_normalisation}, ~\cite{pose_normalisation}. Inspite of this, the real world applications are still not reliable and robust. 

In case of face recognition from video the actual processing was done at frame level which is actually an image. The best frame among all the frames of the video was selected based on quality of the face in the image and recognition algorithm was applied subsequently ~\cite{frame_pose}, ~\cite{frame_image_sets}. With the advent of deep learning architectures, 2D convolution network came to be applied on images or frames of videos to detect and recognise faces. The generation of features was no more manual. Deep features, though undecipherable, were better than manually developed features for face recognition. This lead to increase in accuracy and robustness. However, deep architectures did not treat video as one input rather they also generated spatial features based on series input of frames. 

In the recent works, 3D CNN showed good results for activity recognition in videos \cite{tran_3D_conv}. This is because
unlike 2D CNNs, 3D CNNs are capable of modelling the time dimension as well as the spatial dimension.  The 3D CNNs accepted and treated a video as a single input unit. 3D CNNs could generate a single compact feature that contained facial trait as well as body language, gait pattern and any other temporal and spatial pattern that may be relevant to classification. 

The concept of residual networks \cite{resnet_imagenet} allowed for more depth in deep learning networks without the limitation of vanishing gradient. While 2D residual networks was successfully applied for image classification \cite{resnet_imagenet}, 3D residual networks have been designed to extend the capability of residual networks in third dimension as well \cite{3d_resnet}. 3D residual networks have been successfully applied for activity recognition using videos \cite{3d_resnet}. The application of 3D CNNs and 3D residual networks for activity recognition motivated us to use the 3D CNNs for face recognition using videos. 

Apart from YTF \cite{ytf} dataset, all the other video datasets such as UCF \cite{ucf} and HMDB \cite{hmdb} are available for activity recognition. We have therefore created a comprehensive biometric dataset with modalities  video, iris, and fingerprint. The video dataset has been used in our experiments for face recognition.

In this paper, we perform the face recognition on CVBL facial video dataset. This paper has therefore the following contributions:

\begin{enumerate}
  \item A comprehensive biometric dataset called the CVBL dataset containing video, iris, fingerprint modalities has been collected.
  \item It uses 3D residual networks to find out the accuracy for face recognition in videos.
  \item It compares the accuracy of 3D residual network for different depths of residual networks in case of face recognition in videos.
  \item It also compares the accuracy of different genres of 3D residual networks in case of face recognition in videos.

\end{enumerate}

The previous work on face recognition has been discussed in section \ref{RELATED WORK}.  Section \ref{VIDEO DATASET} discusses in detail about our comprehensive biometric dataset called the CVBL dataset. The residual network architectures used in the experiment and configuration detail related to the experiment have been discussed in section \ref{EXPERIMENTAL CONFIGURATION}. The exact implementation details are discussed in section  \ref{IMPLEMENTATION} followed by discussion over the result in section \ref{RESULTS AND DISCUSSION}. Finally, the conclusion and future scope is discussed in section \ref{CONCLUSION}. 

\section{RELATED WORK} \label{RELATED WORK}

\subsection{Deep architectures for face recognition approaches}

A lot of work has been done in the field of face recognition using images.  Convolution Neural Networks (CNN) are being used for face recognition these days. In \cite{contrastive_loss} the author introduced the concept of contrastive loss. The contrastive loss is based on Euclidean distance between the two points. In contrastive loss, the points in higher dimension are mapped to a manifold such that the euclidean distance between the points on the manifold corresponds to the similarity between the same two points in the higher dimension input space. In contrastive loss, the CNNs are trained using pairs of images. The contrastive loss is such that it tries to generate highly discriminative features when the training images in the pair are dissimilar to each other. In case the images in the pair for training are same, the contrastive loss tries to generate similar features for the images.

In \cite{facenet} triplet loss was introduced. The author trained a CNN using triplets of images containing an anchor image which is the actual image, the positive image which is the image of the same person as in anchor image and a negative image which consists of an image of a person different from that in anchor image. The training is done to obtain discriminative features such that it increases the distance between the anchor face and negative face and decreases the distance between positive and anchor face. In case of both contrastive loss and triplet loss, organising the batches in pair or triplet such that it satisfies a given condition is in itself a difficult and complex process.

In the sequence of improvement of loss function to increase the discriminative power of features for face recognition, a new loss function was proposed by Liu et. al. In his paper \cite{lsoftmax}, Liu proposed a generalized large-margin softmax (L-Softmax) loss which explicitly encourages intra-class compactness and inter-class separability between learned features. L-softmax not only can adjust the desired margin but also can avoid overfitting. 

Liu et. al. \cite{asoftmax} in the year 2017 proposed a new loss function called A-softmax as his extension and improvement to L-softmax. A-Softmax loss can be viewed as imposing discriminative constraints on a hypersphere manifold, which intrinsically matches the prior so that faces also lie on a manifold. The size of angular margin can be quantitatively adjusted by a parameter m. This makes the learning better by increasing the angular margin betweeen the classes and making the feature discrimination better than  L-softmax. This paper has used two datasets for performance analysis. One is Labeled Face in the Wild (LFW) and the other is Youtube Faces (YTF).  A-softmax also which has also been called as SphereFace in the paper, achieves 99.42\% and 95.0\% accuracies on LFW and YTF datasets respectively. In an extension to the angular softmax loss, Deng et. al. in his work \cite{angular_margin} tried to increase the inter-class separability by introducing the concept of additive angular margin. 

Lot of work has been done on images for face recognition. However, all the work on images in face recognition are prone to spoofing. This can be overcome only if we can use videos for face recognition. This will allow the system to check the liveliness of the person by the random body and face movements and thus avoid spoofing.

Activity recognition is one domain where deep learning has been successfully applied for processing temporal domain along with the spatial domain.

\subsection{Activity recognition approaches}

Karpathy et. al. \cite{karpathy} used two-stream convolution network for activity recognition in video. He used one stream to input centrally cropped video frames and other stream to input the full frame but at half the original resolution. The two streams got concatenated later in the fully connected layer. The use of two-stream architecture made the processing of videos 2 to 4 times faster than in case of a single stream architecture. However, the problem of capturing the temporal dimension still remained because the use of 2D convolution in the two-stream architecture limited the architecture from capturing the temporal dimension.

Joe et. al. in \cite{lstm_snippets} applied an array of Long Short-Term Memory (LSTM) cells to capture the temporal dimension in videos for activity recognition. Because of the use of LSTMs, the method was capable of handling full length videos. This meant that the architecture using LSTM was able to model the temporal change across the entire length of the video. Firstly, a layer of CNN  processed frames of videos in sequence to produce spatial features. These spatial features were passed to LSTM for extracting temporal features. Jeff et. al. in his research work \cite{lstm_ltrc} also applied LSTMs in a different architecture but with the same objective of modelling the temporal dimension for activity recognition. However, the LSTM based architectures are not giving accuracy better than two-stream based architectures.

Tran et. al. in his work \cite{tran_3D_conv} used 3D CNNs for activity recognition. Using his 3D CNN, he could capture both spatial and temporal dimension in his features. The features extracted from videos using 3D CNN are highly efficient, compact, and extremely simple to use. He called these features C3D. Tran et. al. demonstrated that C3D features along with a linear classifier can outperform or approach current best methods on different video analysis benchmarks. However, the only problem with 3D CNN is that the 3D CNNs cannot capture the entire length of video sequence in one go. This causes a limitation in capturing the temporal dimension if the length of the temporal activity is longer than the number of frames captured by 3D CNN. 

Tran in his work \cite{tran_3D_conv} used the temporal depth of 16 frames. Laptev et. al. in his work \cite{laptev_3dcnn} tried to figure out what happens to activity recognition accuracy if we change the temporal depth of video clip. Laptev experimented for temporal depth of 16, 20, 40, 60, 80 and 100. He  found that on increasing the temporal depth, the accuracy for activity recognition increased. This was because the 3D CNN architecture could model the activity in a better way when the number of frames were more. Thus, this experiment also confirmed that temporal dimension played an important role in the activity recognition. However, greater temporal depth also required more processing. 

In 2016, He et. al. \cite{resnet_imagenet} came with the idea of  Residual networks and won the first place in several tracks in ILSVRC \& COCO 2015 competitions. However, in ILSVRC the architectures are tested on images.

In 2017, Kensho et. al.  \cite{3d_resnet} extended the concept of residual networks from 2D to 3D. Kensho applied 3D residual networks for activity recognition in videos. He changed the depth of 3D residual networks and tried to experiment its effect on the accuracy. He found that as we increase the depth of the residual networks, the accuracy increases till it reached the depth of 151. Upon further increasing the depth, the accuracy for activity recognition saturates. With this experiment, it was clear that with the increasing depth, it was possible to capture better features and thus increase the accuracy of the activity recognition.

The work of Kensho et. al.  \cite{3d_resnet} motivated us to experiment if the residual networks can be used to identify a person in a video. To realize this purpose, we developed a video dataset of our own for face recognition called the CVBL dataset. 

\section{VIDEO DATASET} \label{VIDEO DATASET}


\begin{figure}[!t]
   \includegraphics[scale=.2]{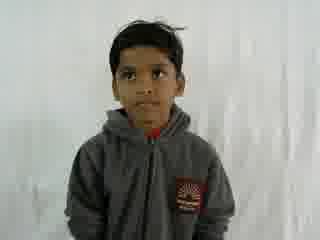}
    \includegraphics[scale=.2]{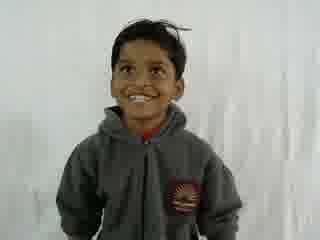}
    \includegraphics[scale=.2]{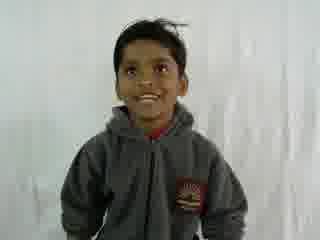}
    \includegraphics[scale=.2]{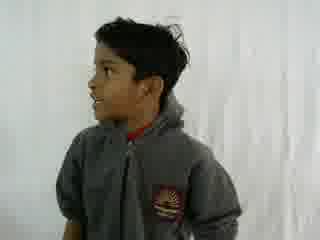}
    \includegraphics[scale=.2]{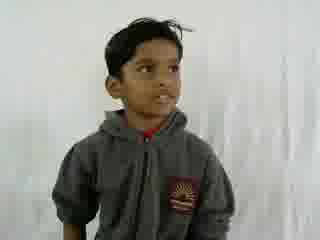}
    \caption{Frames from CVBL dataset}
\label{cvbl_frames}
\end{figure}


CVBL dataset is named CVBL dataset \cite{cvbl} after the lab that is creating it. The CVBL biometric data \cite{cvbl} is an exhaustive biometric dataset consisting of facial videos, fingerprint, and signature of each subject. The dataset consists of biometric data of 125 school going children below the age of 15.

From the CVBL dataset, we used the face video dataset for our face recognition experiment. The face video dataset consists of $320 \times 240$ size video taken at 30 frames per second. Each video is of maximum one minute and minimum five such videos of each subject have been taken. The videos show the subjects talking and expressing themselves freely while being seated at a place in front of the camera as shown in figure \ref{cvbl_frames}. There are 125 different subjects and thus there are 125 classes for face recognition. The subjects are facing the camera however they can move their face in any direction while talking. The videos include static background and there is no camera motion. More number of subjects will be included in the CVBL dataset \cite{cvbl} in the future.

In our experiment, out of total 675 videos, 415 videos have been taken for training and the rest 260 for testing. Thus training is to testing split ratio is 60:40 approx.

\section{EXPERIMENTAL CONFIGURATION} \label{EXPERIMENTAL CONFIGURATION}

\subsection{Summary}

Our objective is to find out the accuracy of 3D ResNets on face recognition video dataset. In addition we also wanted to know how the accuracy of face recognition changes with change in depth and genre of residual networks. For this purpose, we used the code from \cite{3d_resnet} for experimentation and modified it as per our requirements and objectives. The code for face recognition experiments \cite{3d_resnet} uses Pytorch library \cite{pytorch}. We begin our analysis by checking whether the size of the  dataset is large enough not to underfit the residual network of such large depths. We therefore start with the depth of 18, assuming that if ResNet-18 overfits then we can conclude that the size of the dataset is too small to train the architecture of such depth. We will experiment with the larger depths of ResNets only if we are convinced that CVBL dataset is large enough to train ResNet-18 without underfitting.

\subsection{Network architectures}

In this section all those network architectures will be discussed which we plan to implement and analyse them over training them on CVBL dataset. In this paper, ResNet architectures of various depth have been experimented. The ResNet architectures have a special property that they allow shortcut connections to bypass layers in between to move to the next layer. However, back propagation still takes place without any problem. \

Apart from the ResNet (basic and bottleneck blocks) \cite{resnet_imagenet}, following extensions of the ResNet architecture have also been used for experiment: Pre-activation ResNet \cite{pre-activation_resnet}, Wide ResNet (WRN) \cite{wide_resnet}, ResNeXt \cite{resnext}, and DenseNet \cite{densenet}.

A basic ResNet block \cite{resnet_imagenet} is the most simple ResNet and consists of only two convolution layers. Each of the convolution layers is followed by a batch normalization layer and a non-linearisation layer ReLU. A shortcut connection is also provided between the  top of the block and to the layer just before the last ReLU in the block. ResNets-18 and ResNets-34 adopt the basic ResNet block structure.

A ResNet bottleneck block \cite{resnet_imagenet} is different from the basic ResNets block in the sense that it consists of three convolution layers instead of two. As in case of basic ResNets block, each convolution layer is followed by batch normalisation layer and ReLU layer. The first and third convolution layers consists of the filters of size $ 1 \times 1 \times 1 $ whereas the second convolution layer consists of filters of size $ 3 \times 3 \times 3 $. The networks which adopt ResNets bottleneck block are ResNet-50, 101, 152, and 200. The $ 1 \times 1 \times 1 $ convolutions \cite{network} help the network to go deeper by being computationally efficient as well as contains more information than otherwise.

Unlike bottleneck ResNet, where each convolution layer is followed by batch normalization and a ReLU, in case of Pre-activation ResNet \cite{pre-activation_resnet} the batch normalization layer and the ReLU layer come before convolutional layer. 
He et al. \cite{identity_mapping} also confirmed in his studies on ResNet that Pre-activation ResNets are better in optimization and avoiding overfitting. The shortcut in case of Pre-activation ResNets connects the top of the block to  the layer just after the last convolution layer in the block. Pre-activation ResNet-200 is an example using Pre-activation ResNet.\

Wide resnets \cite{wide_resnet} increase the width of the residual network instead of increasing the depth of the residual network. Width here means the number of features maps in one layer. If we talk of a convolution layer network, the number of feature maps corresponds to the number of filters in a convolution layer. In a neural network, the width corresponds to the number of neurons in a layer. In \cite{wide_resnet} the authors increased the width instead of depth and showed that same accuracy can be gained by increasing width instead of depth. Several other authors \cite{wide_resnet} however feel that the increase in accuracy was not because of increase in number of feature maps but because of increase in number of parameters. The increase in number of parameters can also cause overfitting.

Densenets \cite{densenet} are those residual network which exploit the concept of feature resuse. In DenseNets, the features from early layers are used in the later layers by the providing direct connections from every early layer to every later layers in the feed-forward fashion and concatenating them.This process makes the interconnections very dense and hence the name. The concept of pre-activations used in Pre-activation ResNets have also been used in Densenets to reduce the number of parameters and yet achieve better accuracy than ResNets. The number of feature maps at each layer is called the growth rate in case of Densenets. This is because the features maps at a particular layer grows after concatenation with the feature maps of the previous layer. DenseNet-121 and DenseNet-201 with growth rate of 32 are examples of DenseNets.

In \cite{resnext} Xie et al. introduced a new term called cardinality. As per the author, the cardinality refers to the size of the set of transformations. In his paper Xie et. al experimented with 2D residual architectures for image processsing. He showed that increasing the cardinality of 2D architectures is more effective than using wider or deeper ones. With this context, ResNeXt was introduced with the concept of cardinality. Cardinality refers to the number of middle convolutional layer groups in the bottleneck block. These groups divide the feature maps into small groups which are later concatenated. ResNeXt performed the best among all the residual networks in case of activity recognition \cite{3d_resnet}. This showed that increasing the cardinality is better than increasing the width or depth.

\section{IMPLEMENTATION}   \label{IMPLEMENTATION}

\textbf{Training} : For training purpose, a 16-frame clip is generated from the temporal position selected by uniform sampling of the video frames. In case the video contains less than 16 frames, then 16 frames are generated by looping around the existing frames as many times as required. Multiscale cropping is done by first selecting randomly a spatial position  out of 4 corners and 1 center. Then, for a particular sample, a scale value is selected out of the following to perform multi-cropping: $\left\{ \frac{1}{2^{\frac{1}{4}}} ,  \frac{1}{\sqrt{2}} ,  \frac{1}{2^{\frac{3}{4}}},  \frac{1}{2} \right\}$. 

Aspect ratio is maintained to one and scaling is done on the basis of shorter side of the frame. Frames are then resized to $112 \times 112 $ pixels. After all this, we finally get the input sample size as (3 channels $ \times$  16 frames $ \times$  112 pixels $ \times$  112 pixels). Horizontal flipping with a  probability of 50 precent is also performed. Mean subtraction is performed to keep the pixel values zero centered. In the process of mean subtraction, a mean value is subtracted from each color of the sample. Cross-entropy loss is used for calculation of loss and back-propagation. For optimization using the calculated gradients, stochastic gradient descent (SGD) with momentum is used. Weight decay of 0.001 and momentum of 0.9 has been used in the training process. When training the networks from scratch, we start from learning rate of 0.1, and divide it by 10 after the validation loss saturates.
 
Each video is split into non-overlapped 16-frame clips and  each clip is then passed into the network for recognition of faces. Hence in a  way, we are following sliding window to generate input clips where the sliding window is moving in time dimension and the length of the sliding window is 16. The sliding window is being moved in non-overlapped fashion.

\section{RESULTS AND DISCUSSION}  \label{RESULTS AND DISCUSSION}

\begin{table}[!t]
\caption{Accuracy of our proposed method using Residual Networks of different depth on CVBL dataset for face recognition}
\label{accuracy_table_CVBL}       
%
%
\begin{tabular}{p{10cm}p{2cm}}
\hline\noalign{\smallskip}
Residual Networks & Accuracy  \\
\noalign{\smallskip}\svhline\noalign{\smallskip}
\textbf{Residual Networks} & \textbf{Accuracy} \\
ResNet-18  & 96\%   \\
ResNet-34  & 93.7\%   \\
ResNet-50  & 96.2\%   \\
ResNet-101  & 93.4\%   \\
ResNet-152  & 49.1\%   \\
ResNeXt-101  & 78.5\%   \\
Pre-activation ResNet-200  & 96.2\%   \\
Densenet-121  & 55\%   \\
\textbf{Densenet-201}  & \textbf{97\%}   \\
WideResnet-50   & 90.2\%   \\
\noalign{\smallskip}\hline\noalign{\smallskip}
\end{tabular}
\end{table}

\begin{figure}[t]
\includegraphics[width=0.9\linewidth]{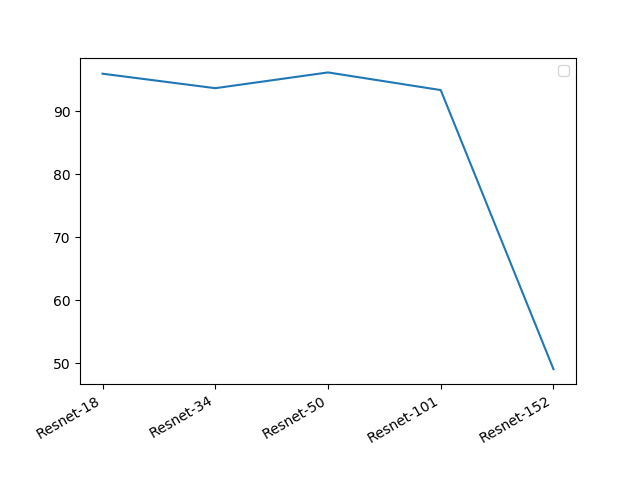} 
\caption{Variation of accuracy of Resnets with depth}
\label{resnet_accuracy_graph}
\end{figure}

\begin{figure}[tp]
 \begin{subfigure}{0.5\textwidth}
\includegraphics[width=0.9\linewidth, height=5cm]{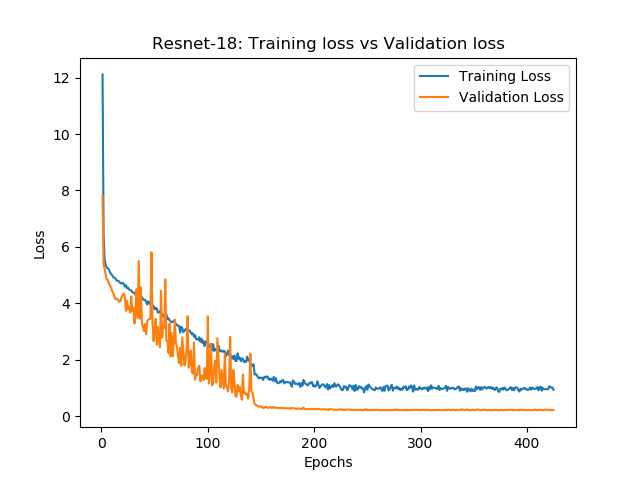} 
\caption{ResNet-18}
\label{resnet-18_train_val_loss}
\end{subfigure}
\begin{subfigure}{0.5\textwidth}
\includegraphics[width=0.9\linewidth, height=5cm]{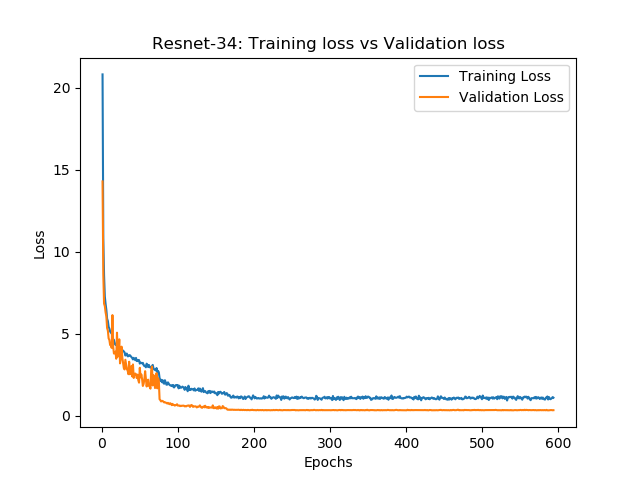}
\caption{ResNet-34}
\label{resnet-34_train_val_loss}
\end{subfigure}
\begin{subfigure}{0.5\textwidth}
\includegraphics[width=0.9\linewidth, height=5cm]{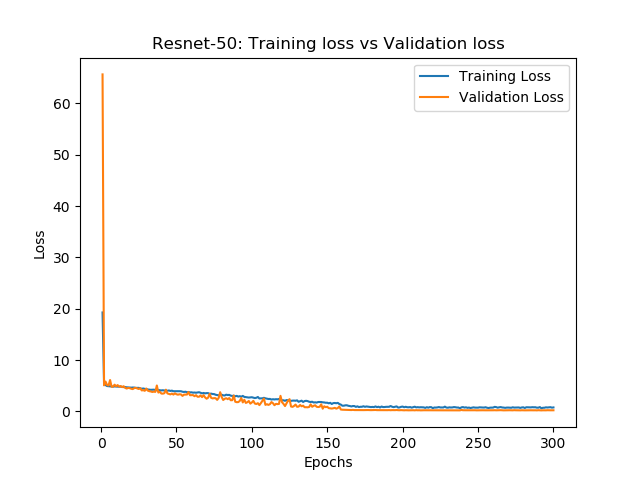} 
\caption{ResNet-50}
\label{resnet-50_train_val_loss}
\end{subfigure}
\begin{subfigure}{0.5\textwidth}
\includegraphics[width=0.9\linewidth, height=5cm]{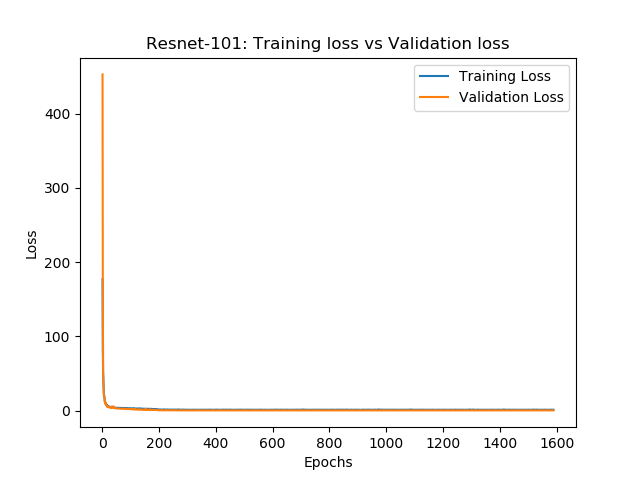}
\caption{ResNet-101}
\label{resnet-101_train_val_loss}
\end{subfigure}
\begin{subfigure}{0.5\textwidth}
\includegraphics[width=0.9\linewidth, height=5cm]{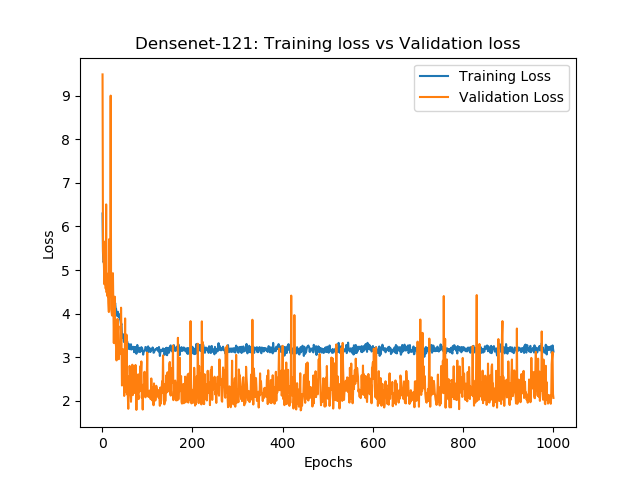} 
\caption{Densenet-121}
\label{densenet-121_train_val_loss}
\end{subfigure}
\begin{subfigure}{0.5\textwidth}
\includegraphics[width=0.9\linewidth, height=5cm]{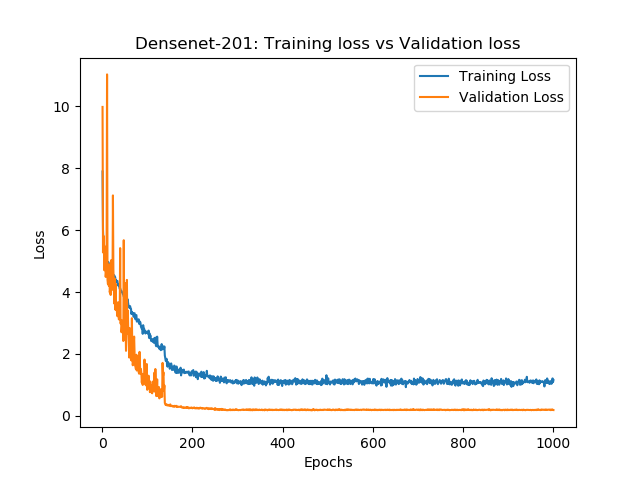}
\caption{Densenet-201}
\label{densenet-201_train_val_loss}
\end{subfigure}
\caption{Training loss vs validation loss for different Resnets}
\label{train_val_loss_1}
\end{figure}

\begin{figure}[t!]
\begin{subfigure}{0.5\textwidth}
\includegraphics[width=0.9\linewidth, height=5cm]{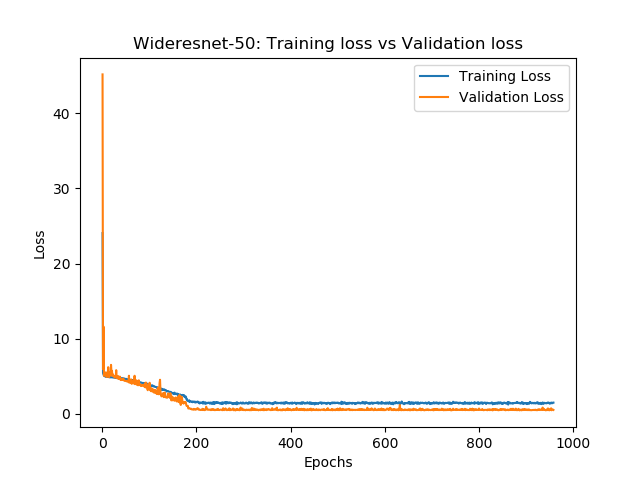} 
\caption{Wideresnet-50}
\label{wideresnet-50_train_val_loss}
\end{subfigure}
\begin{subfigure}{0.5\textwidth}
\includegraphics[width=0.9\linewidth, height=5cm]{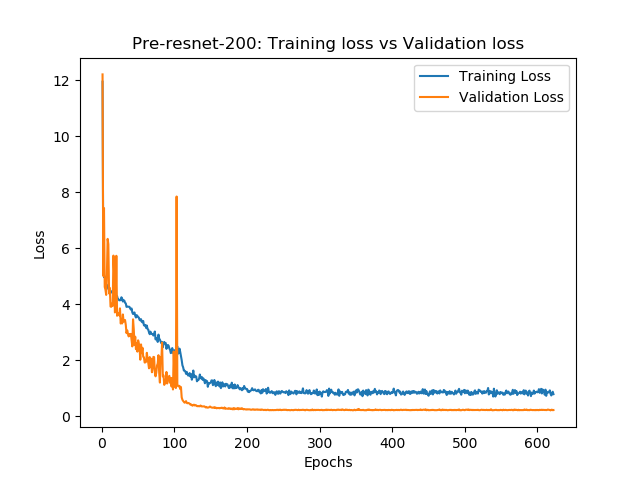}
\caption{Pre-activation ResNet-200}
\label{pre-resnet-200_train_val_loss}
\end{subfigure}
\begin{subfigure}{0.5\textwidth}
\includegraphics[width=0.9\linewidth, height=5cm]{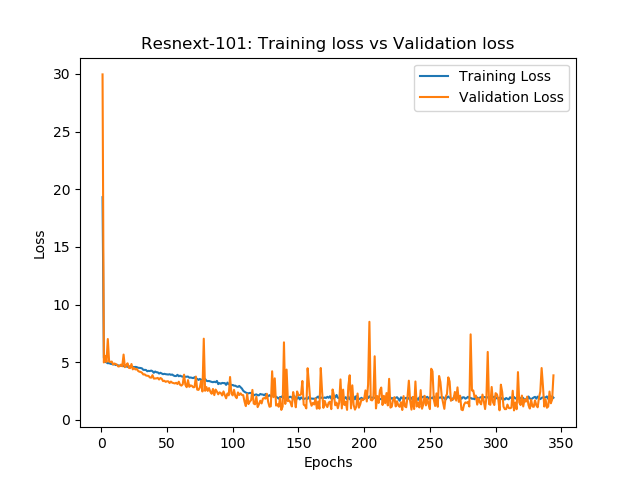} 
\caption{ResNeXt-101}
\label{resnext-101_train_val_loss}
\end{subfigure}
\begin{subfigure}{0.5\textwidth}
\includegraphics[width=0.9\linewidth, height=5cm]{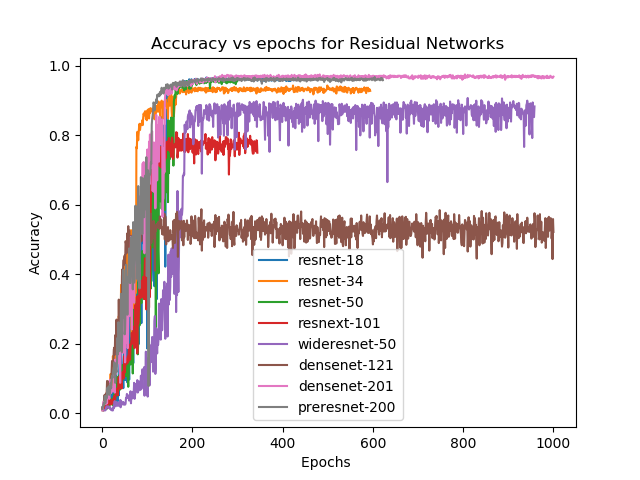}
\caption{Accuracy in all kinds of ResNets of different depth}
\label{accuracy_all}
\end{subfigure}
\caption{Figure a,b and c show training vs validation loss for Wideresnet-50, Pre-activation ResNet-200 and ResNeXt-101.\\
Figure d shows the graph of accuracy against the number of epochs for all residual networks }
\label{train_val_loss_2}
\end{figure}

As we discuss the results, it must be noted here that the term ResNets means ResNet-18, 34, 50, 101 and 152. Extension of ResNets means Pre-activation ResNets, Wide Resnets and DenseNets. The results of experiments on CVBL dataset are summarized in table \ref{accuracy_table_CVBL}. The training vs validation loss  and comparison of accuracy for all kinds of ResNet architectures are shown in figures \ref{train_val_loss_1} and \ref{train_val_loss_2}.

It can be easily observed that in case of ResNets, the accuracy does not increase with the increase in depth of the architecture. Infact it follows a zig-zag kind of path  as shown in figure \ref{resnet_accuracy_graph}. For ResNet-18 the accuracy is 96\%. It drops to 93.7\% for ResNet-34. The accuracy of ResNet-50 then comes back to 96.2\% and then dropping to accuracy of 93.4\% for ResNet-101. Thus there is a zig-zag kind of variation in accuracy as we increase the depth of the ResNets. It is also observed from the table \ref{accuracy_table_CVBL} and figure \ref{resnet_accuracy_graph} that the performance of ResNets drops very sharply after ResNet-101. ResNet-152 fails to perform with just 49.1\%. The increase in depth for the same training set may be the cause for the high bias and hence the decrease in accuracy. Inspite of all these variations of accuracy in ResNet architectures, it can be inferred that ResNets in general are performing well with more than 90\% accuracy. The highest accuracy in ResNets has gone to 96.2\%. 

In case of Densenets too, DenseNet-201 performed the best with 97\% accuracy. The resuse of features from previous layers in later layers of Densenet seems to have contributed to increase in the accuracy to 97\% in case of Densenet-201. However, the accuracy of DenseNet-121 was way too low with just 55\%. Figure \ref{densenet-121_train_val_loss} and \ref{densenet-201_train_val_loss} show the training vs validation loss for Densenet-121 and Densenet-201 respectively. It can be seen that after convergence the training loss in case of Densenet-121 is more than that in case of Densenet-201. This shows that Densenet-121 has a high bias as compared to Densenet-201. Also Densenet accumulates features from previous layers to later layers. Hence we can say that Densenet-201 is able to make use of the accumulated features because of its high depth. Due to its relatively lower depth, Densenet-121 is not able to make use of the accumulated features to increase its accuracy.

Wide ResNet-50 gave an accuracy of 90.2\%. ResNet-50 which is of the same depth as wideresnet-50 gave an accuracy of 96.2\%. Wideresnet-50 not only failed to maintain the accuracy to 96.2\% but made the accuracy lower. Thus, it is evident that in case of face recognition from videos, increasing the width of the architecture only had a negative effect on the accuracy. 

Pre-activation ResNet also performed equivalent to the best performers with an accuracy of 96.2\%. In Pre-activation ResNets, the ReLU and batch normalization layer are placed before the convolution layer unlike in Resnets. Pre-activation ResNets are said to perform better than Pre-activation ResNets \cite{3d_resnet} because of change of the sequence of layers. Interestingly, in our case, change in sequence of layers had no effect on the accuracy. For the sake of comparison, the rise in accuracy of all kinds of resnets with respect to epochs is shown in figure \ref{accuracy_all}.

Our experiments conducted to understand the behavior of residual networks for face recognition on CVBL video dataset found the results highly encouraging with the highest accuracy going up to 97\%. We can safely interpret that residual networks are sensitive to face recognition and obtaining good results on CVBL video dataset.

\subsection{Comparison with state-of-the-art in face recognition}
\begin{table}[!t]
\caption{State-of-the-art results for face recognition on image and video datasets.}
\label{accuracy_table_all}       
\begin{tabular}{p{4cm}p{2cm}p{2cm}p{2cm}p{2cm}}
\hline\noalign{\smallskip}
\textbf{Architecture} & \textbf{Image dataset} & \textbf{Accuracy} & \textbf{Video dataset} & \textbf{Accuracy} \\
\noalign{\smallskip}\svhline\noalign{\smallskip}
Facenet\cite{facenet}              & LWF                    &  99.63\%          &        YTF              &  95.12\%         \\
Deep ID 2\cite{deepid2}            &  --                    &    --             &        YTF                &    93.2\%         \\
Center Loss\cite{center_loss}          &  LWF                    &    99.28\%             &        YTF          &  94.9\%         \\
Deep Residual 
Learning                    &                       &                       &                    &              \\
on Images \cite{resnet_imagenet}                          &  Imagenet             &    3.57\% error            &     --             &   --         \\
L-softmax\cite{lsoftmax}              &  LWF             &         98.71\%            &     --                 &   --         \\
A-softmax \cite{asoftmax}             &  LWF             &         99.42\%            &     YTF               &   95.0\%         \\
3D Residual networks                  &                   &                              &                  &                  \\
(ResNeXt-101 64 frames)\cite{3d_resnet}  &  --             &         --                  &   UCF-101         &   94.5\%           \\
3D Residual networks                    &                 &                              &                    &                   \\
(ResNeXt-101 64 frames)\cite{3d_resnet}   &  --             &         --                  &   HMDB-51         &   70.2\%            \\
\noalign{\smallskip}\hline\noalign{\smallskip}
\end{tabular}
\end{table}

We compare table \ref{accuracy_table_CVBL} and table \ref{accuracy_table_all} for comparing the results of our proposed method with the state-of-the-art results on various image and video datasets. Table \ref{accuracy_table_all} shows the accuracy of different approaches on LWF dataset for face recognition in images and YTF dataset for face recognition in video. From Table \ref{accuracy_table_all} it is observed that for face recognition in images, the accuracy of 99\% has been achieved by different approaches. However, in case of face recognition in videos our approach using  DenseNet-201 has successfully achieved the accuracy of 97\%. This accuracy is well above the state-of-art accuracy of 95.12\% by nearly 2\%.

\section{CONCLUSION}  \label{CONCLUSION}

ResNets architectures seem to be sensitive to face patterns in videos. Since the CVBL dataset consists of videos with same background which is plain white, it can be easily assumed that the background is not at all contributing to the recognition accuracy. It is only the spatial and the temporal dimensions that are contributing effectively to the classification and recognition accuracy. Except few cases, the ResNets are providing good results with accuracies above 90\%. DenseNets performed the best with 97\%. 

Hence we can conclude that ResNets are sensitive to face recognition patterns with accuracy near 96\%. This is the first of its kind experiment on face video dataset and the residual networks have given an accuracy of above 90\% in general which gives a very positive indication about the future in video biometric.

In future we plan to collect biometric samples from more subjects and prepare a bigger biometric and exhaustive dataset. We then plan to experiment on this bigger biometric dataset to evaluate the effect of large number of classes in the dataset on face recognition accuracy.  We also plan to experiment on the existing YTF dataset using the residual networks.

\end{document}